\documentclass{article}
\usepackage{ijcai26}

\usepackage{times}
\usepackage{soul}
\usepackage{url}
\usepackage[hidelinks]{hyperref}
\usepackage[utf8]{inputenc}
\usepackage[small]{caption}
\usepackage{graphicx}
\usepackage{amsmath}
\usepackage{amsthm}
\usepackage{booktabs}
\usepackage{algorithm}
\usepackage{algorithmic}
\usepackage[switch]{lineno}

\usepackage{amssymb}

\title{Transformer Interpretability from Perspective of Attention and Gradient}

\author{
Yongjin Cui$^1$
\and
Xiaohui Fan$^1$\and
Huajun Chen$^{1}$
\affiliations
$^1$Zhejiang University\\
\emails
\{cuiyongjin, fanxh, huajunsir\}@zju.edu.cn
}

\begin{document}

\maketitle

\begin{abstract}
Although researchers' attention is more focused on the performance of Transformer models, the interpretation of Transformer can never be ignored. Gradient is widely utilized in Transformer interpretation. From the perspective of attention and gradient, we conduct an in-depth study of Transformer interpretation and propose a method to achieve it by guiding the gradient direction—or more precisely, the attention direction. The method enables more comprehensive interpretation of feature regions, offers detail interpretation, and helps to better understand Transformer mechanism. Leveraging the difference in how Vision Transformer (ViT) and humans perceive images, we alter the class of an image in a way that is almost imperceptible to the human eye. This class rewriting phenomenon may potentially pose security risks in certain scenarios.

\end{abstract}

\section{Introduction}\label{introduction}
\begin{figure*}[]
    \centering
    \includegraphics[width=1.9\columnwidth]{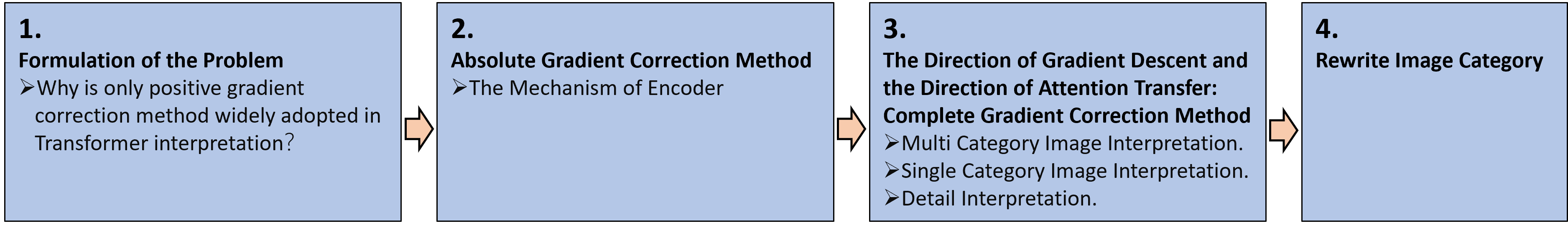} 
    \vspace{-0.3cm}
    \caption{The overall idea of this work.}
    \label{silu}
    \vspace{-0.5cm}
    \end{figure*}

Transformer was first proposed and applied in the field of natural language processing \cite{1}, and later successfully applied to computer vision \cite{6,7} as well as audio field \cite{12,13}, multimodal field \cite{21,22,23} and other crossing fields like pharmaceutical \cite{28,29,30}. And it has spawned a series of large models \cite{38,41,42}, promoting the further development of artificial intelligence. The proposal of Transformer is a major turning point in the development of artificial intelligence, and its excellent performance further drives people to interpret its internal mechanisms.

With the significant development of artificial intelligence driven by Transformer, people's concerns have arisen. These concerns mainly stem from the unexplainable characteristics of AI, which may bring unknown problems, such as algorithmic safety issues in AI healthcare and autonomous driving, and algorithmic fairness issues in treating different groups. Therefore, relevant regulations \cite{43,44} were born. 

As Transformer gives rise to various large models, the contradiction between AI, computing power, and energy is intensifying. At present, the testing method for model performance is still based on the performance of the large test set. If we can accurately verify whether the model has learned the correct knowledge in a timely manner from the perspective of model interpretation, and guide model training based on the interpretation results, it is possible to save more computing power and energy, thereby alleviating the conflict and contradiction.

In summary, we can summarize the goals we hope to achieve through model interpretation: firstly, to understand the internal mechanism of the model; secondly, to understand the basis for models to make inferences; thirdly, to directly test whether the model has learned real knowledge, rather than just judging through indirect indicators such as correct recognition rate.

In this paper, we conduct an in-depth study of Transformer interpretation from perspective of attention and gradient, and propose a method to achieve Transformer interpretation by guiding the gradient direction or, more precisely, the attention direction. The method enables more comprehensive interpretation of feature regions, offers detail interpretation, and helps to better understand Transformer mechanism. During our exploration of pixel-level Transformer interpretation, leveraging properties of gradient, we alter the class of an image in a way that is almost imperceptible to the human eye. This phenomenon may potentially pose security risks in certain scenarios.

\textbf{The overall idea of this work is shown in Figure \ref{silu}.}

\textbf{Our contributions mainly consist of the following five aspects:}
\begin{itemize}
    \item By interpreting the phenomenon of attention transfer, we explained why previous researchers only adopted the positive gradient correction scheme while abandoning the complete and absolute gradient correction schemes.
    \item We have elucidated the mechanism of the Encoder through the absolute gradient correction scheme.
    \item The complete gradient correction scheme performs better in multi-class image tasks. By employing external attention guidance (or, in other words, transforming single-class images into multi-class images), we achieve superior interpretation for single-class images.
    \item We have, for the first time, achieved detail interpretation through internal attention guidance and applied it in practice, elucidating the true causes of the cases presented in the TIS \cite{DBLP:conf/iccvw/EnglebertSNMSCV23} paper, refer to Figures \ref{n7} and \ref{xijiebuchong}.
    \item We leveraged the differences in how humans and ViT perceive images to rewrite the image content in a manner that is imperceptible to humans.
\end{itemize}

\section{Related work}

The interpretation methods of the Transformer model mainly utilize its unique attention structure \cite{1,6}, or combine with the interpretation methods such as GradCAM \cite{51}, LayerCAM \cite{52}, LRP \cite{53}, etc. Vaswani et al. \cite{1} applied the attention of partial layers and partial heads to interpret the intrinsic mechanism of Transformer when they first proposed it, and found that different heads perform different tasks. Multi-head mechanism has become a very important issue in model interpretation. Michel et al. \cite{57} obtained the same conclusion that different heads perform different tasks and contributed differently, and therefore proposed that pruning the unimportant heads has little impact on the model. Considering information originating from different tokens gets increasingly mixed, making attention weights unreliable as interpretation probes, Abnar et al. \cite{58} proposed attention rollout and attention flow to quantify the flow of information through self-attention. And Dosovitskiy et al. \cite{6} applied attention rollout to compute maps of the attention from the output token to the input space when they first proposed Vision Transformer (ViT). Chefer et al. \cite{49} introduced Transformer Attribution (T-Attr) integrating scores throughout the attention graph, by incorporating both LRP-based relevancy and gradient information, in a way that iteratively removes the negative contributions. Chefer et al. \cite{48} introduced Generic Attention-model Explainability (GAE), which combining gradient with multi-head attention maps, and then performing attention rollout. Yuan et al. \cite{yuan2021explaining} interpret information flow inside Vision Transformers using Markov Chain (TAM). Barkan et al. \cite{47} propose Deep Integrated Explanations (DIX), generates explanation maps by integrating information from the intermediate representations of the model, coupled with their corresponding gradients. Xie et al. \cite{DBLP:conf/ijcai/Xie0CZ23} propose ViT-CX based on patch embeddings, rather than attentions paid to them, and their causal impacts on the model output. Englebert et al. \cite{DBLP:conf/iccvw/EnglebertSNMSCV23} propose Transformer Input Sampling (TIS) a perturbation-based explainability method for Vision Transformers, which computes a saliency map based on perturbations induced by a sampling of the input tokens. Chen et al. \cite{46} propose Beyond Intuition Method (BI) to approximate token contributions inside Transformers, in order to solve  the ambiguity of the expression formulation which can lead to an accumulation of error. Zhao et al. \cite{DBLP:conf/icml/ZhaoWZZC24} propose Grad-ECLIP to interpret Contrastive Language-Image Pre-training (CLIP).

Since GAE first introduced gradients into the interpretation of Transformer, subsequent methods such as DIX, BI, Grad-ECLIP, TAM, and others have also incorporated gradients into their frameworks.

\section{Method}\label{sec_method}

We believe that the main reason why neural networks are difficult to interpret is the existence of nonlinear operations. The operation of the attention map $A$ is linear, which is the most distinctive structure in Transformer. The attention map targets patch information rather than individual pixels. Therefore, model interpretation methods based on the attention map are all at the patch level. Inspired by attention rollout, we attempted to interpret the model using the linear operation part. Attention rollout is based on the assumption that the identities of input tokens are linearly combined through the layers based on the attention weights. In order to convert multi head attention into single head attention, attention rollout utilizes an equal weighted average approach for different heads. An identity matrix $I$ is added to represent the contribution of the residual connection part, and then the rollout is executed. It is a linear approximation of Transformer shown as below.
\begin{equation}\label{eq1}
    \bar{A}^{(l)}=I+\mathbb{E}_HA^{(l)}
\end{equation}
\begin{equation}\label{eq2}
    rollout=\bar{A}^{(L)}\cdot\bar{A}^{(L-1)}\cdot...\cdot\bar{A}^{(2)}\cdot\bar{A}^{(1)}
\end{equation}
\begin{equation}\label{eq3}
    C=rollout[0,1:]
\end{equation}
Where $H$ represents number of heads, $L$ represents number of attention layers, $l$ represents the $l_{th}$ layer, $L_{th}$ layer represents the last layer. Because the final input to the MLP head is the $cls$ token, attention rollout takes the $rollout$ value of $cls$ token as $C$, the overall contribution value from the perspective of input or overall attention value from the perspective of model. 

Due to the fact that the input dimension of the final MLP head is the patch embedding dimension, but model interpretation methods based on attention maps take patch information as the minimum unit, and MLP head is also a nonlinear operation, attention rollout does not consider the role of MLP head, in other words, it can only interpret the features extracted by Transformer encoder.

Gradient is often used for parameter optimization in model training, but it itself is an indicator of the relative importance of parameters to the loss. Gradients of model parameters, intermediate values, and other data can serve as indicators to reflect their impact on the model loss to a certain extent. The gradient calculation process covers all subsequent data processing including nonlinear processes in Transformer encoder and the MLP head which attention rollout overlooks. Gradient of certain value is its influence, or contribution on the loss. From another perspective, the gradient of the current value is the weight of the subsequent data processing on the current value. 

\begin{figure}[]
    \begin{center}
    \centerline{\includegraphics[width=0.8\columnwidth]{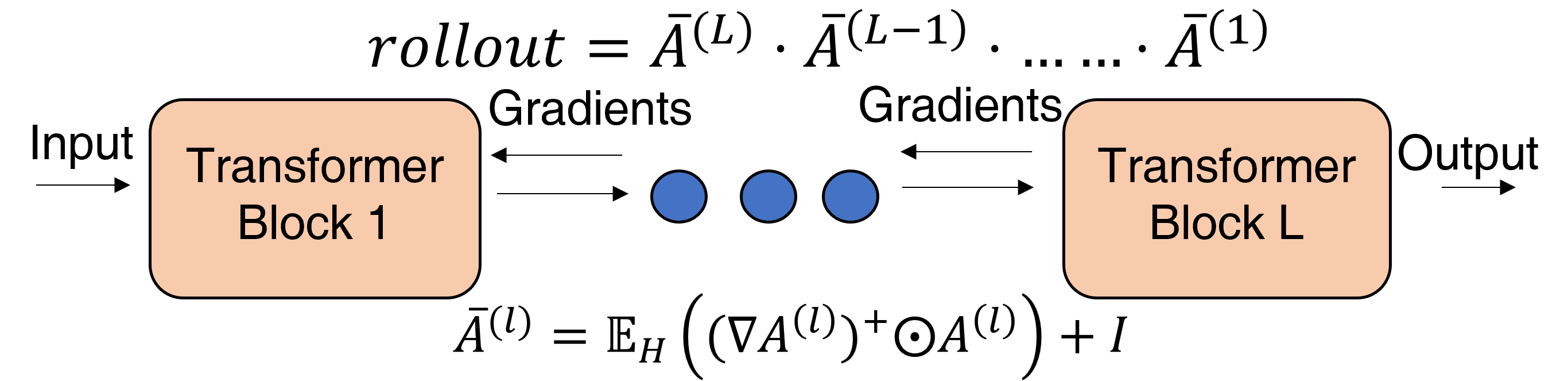}}
    \vspace{-3mm}
    \caption{GAE}
    \label{gae}
    \end{center}
    \vspace{-0.8cm}
    \end{figure}

We believe that Transformer interpretation should leverage the attention mechanism and approximate its computational processes as much as possible, and finally take the attention distribution as the interpretation result. The computational framework of GAE aligns with this approach, although the original authors proposed it from a \textbf{correlation} perspective. The GAE method is shown in Figure \ref{gae}, and the $\bar{A}^{(l)}$ computational process is as follows:
\begin{equation}\label{eq4}
    \bar{A}^{(l)}=I+\mathbb{E}_H((\nabla A^{(l)})^{(+)}\odot A^{(l)})
\end{equation}
Where the gradient is calculated by setting loss as the logit of the target category. And all the negative gradients are set to zero.

The gradient is a vector with two attributes: one is direction, and the other is magnitude. The positive or negative direction indicates whether the variable's contribution to the loss is positive or negative, while the magnitude represents the degree of influence its change has on the result. Existing methods calculate gradient solely based on the logit of the target class and then retain only the positive gradient information.

Our main approach is to set the weights of the attention map separately to the complete gradient (Equation \ref{eq5}) and the absolute gradient (Equation \ref{eq6}). The complete gradient correction is aimed at ensuring the preservation of as much information as possible and achieving a higher degree of approximation to the model. The absolute gradient correction is intended to eliminate the influence of the classifier and focus solely on the attention behavior of Transformer.
\begin{equation}\label{eq5}
    \bar{A}^{(l)}=I+\mathbb{E}_H((\nabla A^{(l)})\odot A^{(l)})
\end{equation}
\begin{equation}\label{eq6}
    \bar{A}^{(l)}=I+\mathbb{E}_H(\left|\nabla A^{(l)}\right|\odot A^{(l)})
\end{equation}
The setting of the loss function has been overlooked by other researchers, yet it is another crucial aspect apart from the gradient. Current methods typically set the loss as the logit of the target class (\ref{eq7}). In reality, the setup of the loss function is pivotal in steering the gradient direction, or rather, the direction of the model's attention transfer. We use another type of loss (\ref{eq8}) in some experiments. We will elaborate on this through experiments in the following sections. \textbf{In subsequent experiments, unless otherwise specified, the loss function will adopt Equation \ref{eq7}.}
\begin{equation}\label{eq7}
    loss=logit_{target}
\end{equation}
\begin{equation}\label{eq8}
    loss=(logit_{class1}-logit_{class2})/logit_{class1}
\end{equation}

\section{Experiment and Results}\label{experiment}

All our experiments are conducted on a NVIDIA A100 GPU. Unless otherwise specified, the loss function will adopt Equation \ref{eq7}. Unless otherwise specified, the Transformer model is ViT\_base \cite{6}.Except for Section \ref{encoder}, when we refer to our method, we are implying complete gradient correction.
\subsection{Formulation of the problem}
In methods involving attention or gradients, GAE is the earliest proposed and widely accepted framework, as exemplified by subsequent methods like DIX that incorporated GAE as part of their computational processes. GAE has evolved into a foundational architecture. The critical question prompting this study is why GAE schemes adopting only the positive gradient correction approach gains broad recognition and application, while alternatives like the complete gradient scheme and absolute correction scheme are abandoned. This paper initiates a series of investigations stemming from this question. To control experimental variables, the baseline for comparison in our analyses is GAE.

\subsection{Why is only positive gradient widely adopted?}

We attempt to understand why researchers have adopted the use of positive gradient for Transformer interpretation. The concept of using gradient to correct the attention map was first introduced by GAE. In GAE, the description is, "Following [5], we remove the negative contributions before averaging" ([5] refers to T-Atrr). In T-Attr, the concept of using gradient is described as, "In order to compute the weighted attention relevance, we consider only the positive values of the gradients-relevance multiplication, resembling positive relevance". This might be the most straightforward way to highlight the regions relevant to the target class.

Subsequent methods also adopted the positive gradient. This may have become an inherent concept. We temporarily call these methods as the reductionist methods, which focus on the interpretation of each component and only retain positive contributions. This is contrary to our viewpoint we introduced in Section \ref{sec_method}, where we believe that we should try to accurately approximate the calculation process of attention, retain more attention information, and obtain better interpretation results through the collaborative effects of each component instead of focusing on individual components separately. We temporarily refer to our approach (Equation \ref{eq5} and \ref{eq6}) as holistic method.

\begin{figure}[]
    \begin{center}
    \centerline{\includegraphics[width=0.8\columnwidth]{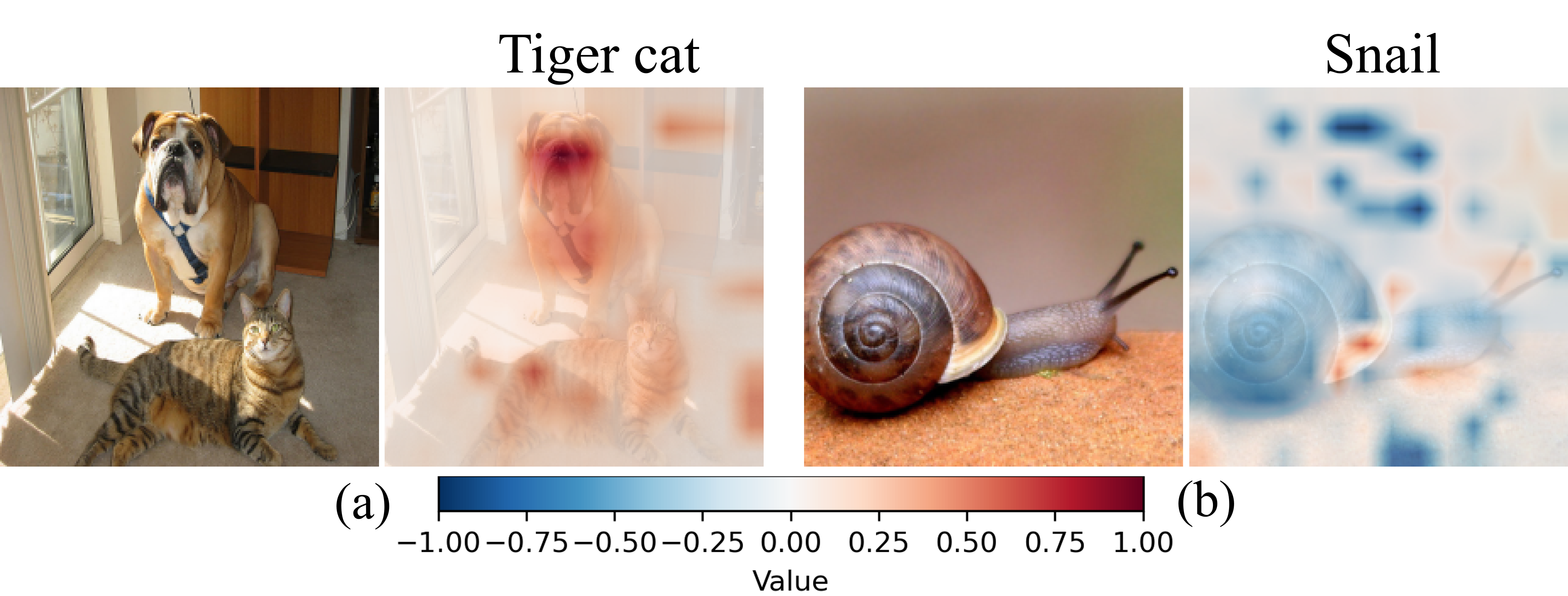}}
    \vspace{-3mm}
    \caption{Experimental results of absolute gradient correction (a) and  complete gradient correction (b). (Absolute gradient correction lacks the ability to distinguish between classes, while complete gradient correction cannot fully interpret the feature region.)}
    \label{n1}
    \end{center}
    \vspace{-0.8cm}
    \end{figure}

Another possible reason that might lead other researchers to solely adopt positive gradients for attention map correction could be the poor experimental results of complete gradient correction and absolute gradient correction (shown in Figure \ref{n1}).

We denote the attention scores as $S$, and then normalize $S$ using $\frac{S}{max(abs(S))}$, which can maintain a linear relationship between scores and preserve their positive and negative attributes. As shown in Figure \ref{n1}, in order to better demonstrate the allocation of overall attention, we use the attention heatmap as the bottom layer, where red represents positive attention, blue represents negative attention, and white represents no attention, and then overlay the original image with a certain transparency on bottom layer for indication. A color bar is used for matching color and value. Colors of an image represent relative value within a single image, and cannot be used for comparing numerical values between images. Other images in this article follow these rules.

\subsection{The Mechanism of Encoder}\label{encoder}
\begin{figure}[]
    \begin{center}
    \centerline{\includegraphics[width=0.7\columnwidth]{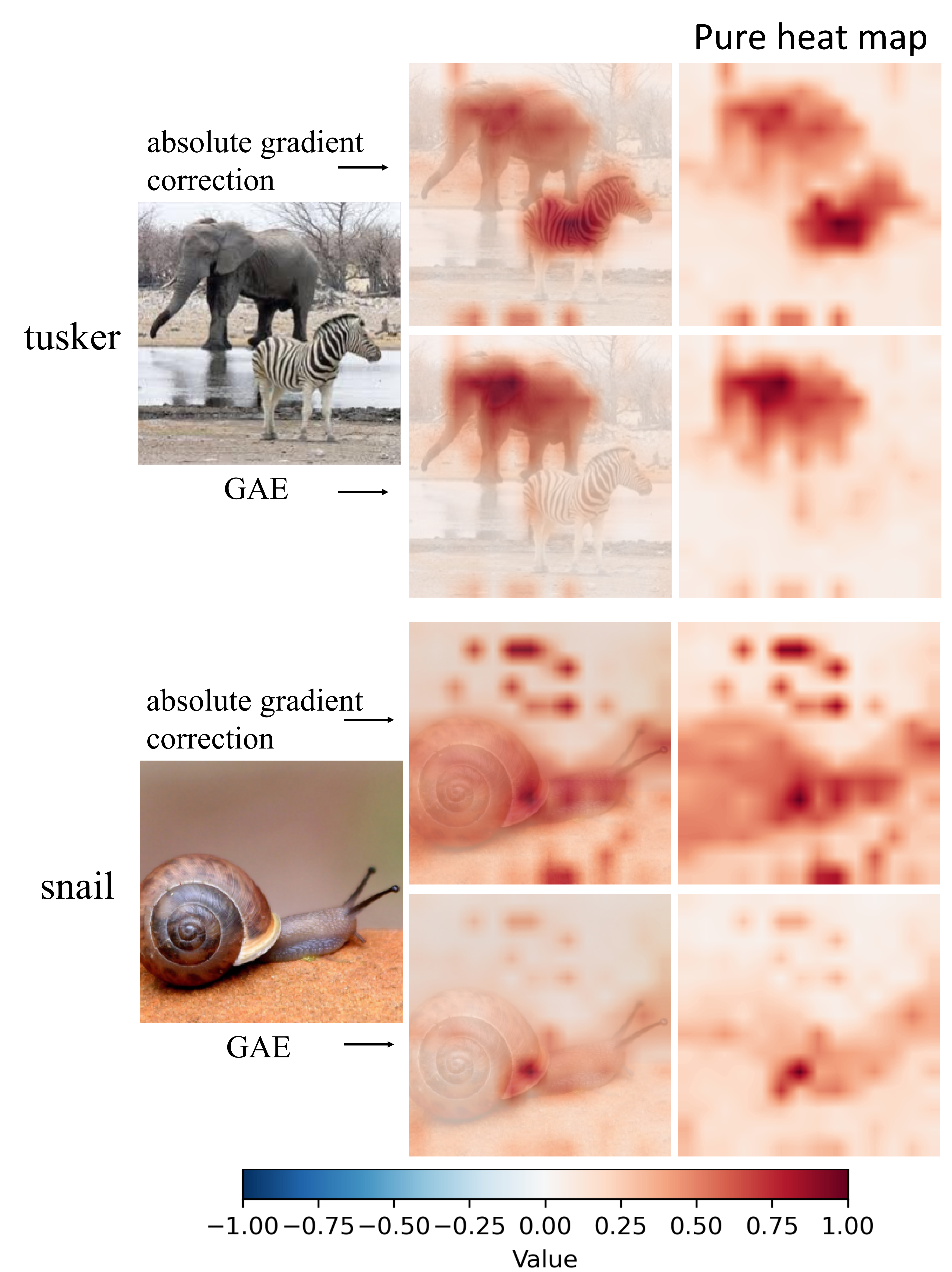}}
    \vspace{-3mm}
    \caption{Absolute gradient correction and positive gradient correction (GAE) interpretation results. The class to be interpreted of the left is tusker, and the right is snail.(Absolute gradient correction method cannot distinguish categories.)}
    \label{n2}
    \end{center}
    \vspace{-0.8cm}
    \end{figure}

When using absolute gradient to correct the attention map, the interpretation results only reflect the model's attention patterns eliminating the influence of the classifier. We can understand the model's mechanism through these attention patterns. Transformer merely extracts features of the learned classes while ignoring the background, as shown in Figure \ref{n1} (a) and Figure \ref{n2}.

\subsection{The Direction of Gradient Descent and the Direction of Attention Transfer}\label{direction}
When interpreting the model, if the loss is set as the logit of the target class. The way to achieve this goal, in terms of adjusting the model parameters, manifests as the parameters being adjusted in the direction of gradient descent. In terms of image content, this is reflected as attention shifting away from the characteristic regions of the target class. The path along which attention shifts the fastest is from the target feature regions to the feature regions of other classes. For example, in Figure \ref{n3}, attention starts from the red area and turns to the blue area. The method we employ to achieve the fastest attention transfer is by correcting the attention map using its gradients. The direction of gradient descent and the direction of attention transfer are two expressions that describe the optimal way to reduce the loss from the perspectives of model parameters and image content, respectively. We will not strictly distinguish between them in the subsequent text.

From another perspective, the positive and negative attributes of the gradient reflect whether the current variable is making a positive or negative contribution to the loss. The positive and negative of the gradient-corrected attention also ultimately reflects whether the feature region is making a positive (red region) or negative (blue region) contribution to the final classification judgment.

\subsubsection{Multi Category Image Interpretation}
\begin{figure}[]
    \begin{center}
    \centerline{\includegraphics[width=0.7\columnwidth]{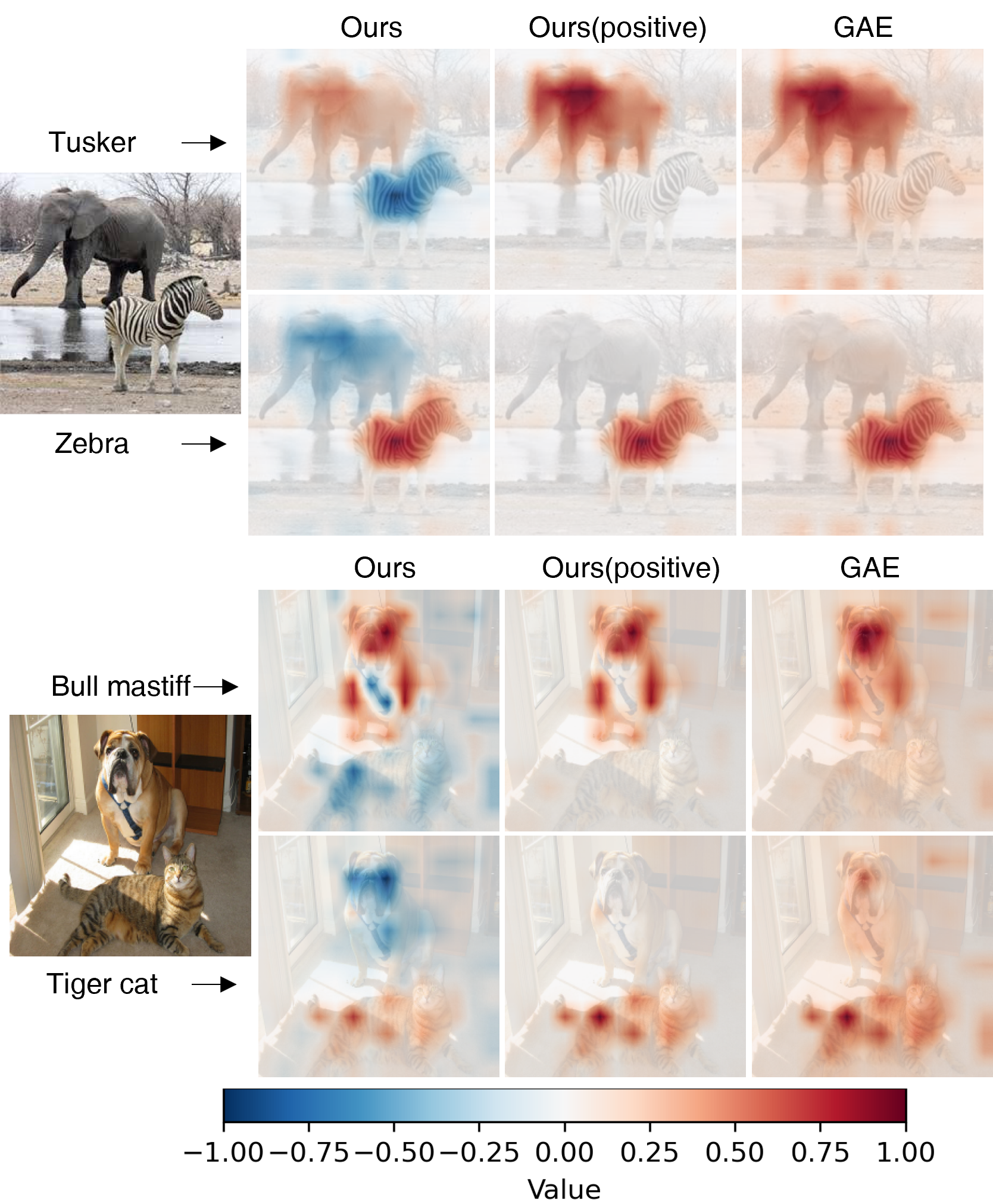}}
    \vspace{-3mm}
    \caption{Interpretation results of ViT model in multi category image recognition task.(Our method can simultaneously allocate both positive and negative attention with less noise.)}
    \label{n3}
    \end{center}
    \vspace{-0.8cm}
    \end{figure}
In multi-class image interpretation, the complete gradient correction method can simultaneously interpret the target class region as well as other classe regions, as illustrated in Figure \ref{n3}. By comparison, it can be observed that the interpretation results of the complete gradient correction method are more accurate and have less background noise compared to GAE that only use positive gradient for interpretation. 

From a computational perspective, complete gradient correction provides a more comprehensive approximation of the model's computational process. The involvement of negative gradients offers greater flexibility in the computational process, correcting some errors and leading to better interpretation results.

The advantages of complete gradient correction in multi-class image interpretation are as follows: \textbf{(1)} It can simultaneously allocate both positive and negative attention, providing a more comprehensive interpretation. This may be particularly useful in other applications, such as when genes with antagonistic relationships jointly determine an indicator. \textbf{(2)} Due to the involvement of negative gradients, it more fully approximates the model's process, resulting in interpretation outcomes with less noise. \textbf{(3)} This approach can distinguish the target class entirely based on the positive and negative attributes of attention. For instance, in the interpretation results of GAE, some degree of positive attention is often allocated to regions of other classes. However, our method allocates negative attention to these regions, resulting in more accurate interpretations. This may be of significant importance in fields that demand higher precision.

\begin{figure}[]
    \begin{center}
    \centerline{\includegraphics[width=1\columnwidth]{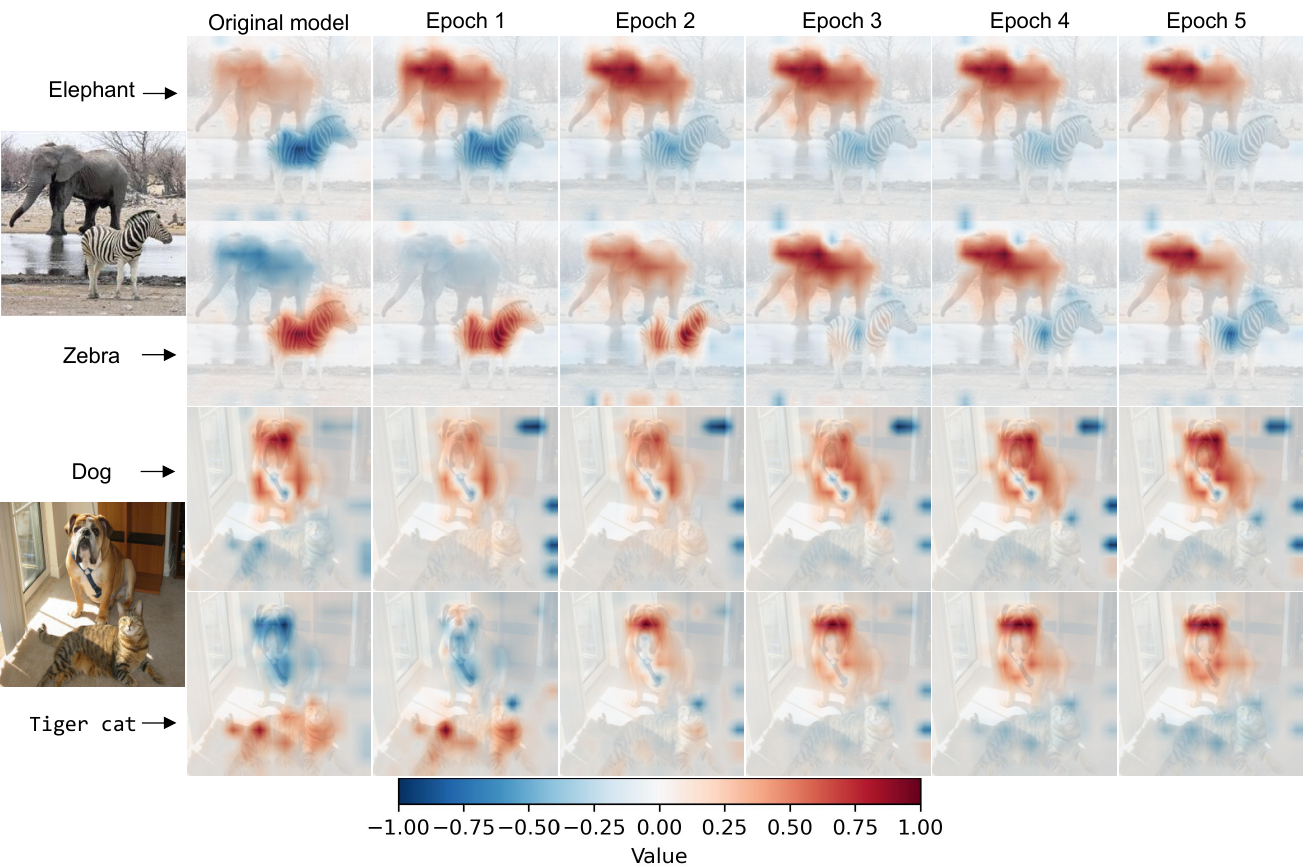}}
    \vspace{-3mm}
    \caption{Attention transfer experiment. (The logit of zebra and tiger cat are used as the loss, and the learning rate is 0.0004.)}
    \label{n4}
    \end{center}
    \vspace{-0.8cm}
    \end{figure}

To validate our interpretations regarding the direction of gradient descent and the direction of attention transfer, we designed the experiment in Figure \ref{n4}. We treat attention maps as the only parameter to be optimized, and refer to the category to be interpreted in the multi class image as the target category, and the other category as the opposing category. In the experiment, the logit of the target category is used as the loss. Each updated model is interpreted using our method. Although fixing attention maps as the only parameter to be optimized may have some impact on gradient calculation, the interpretation results using our method in Figure \ref{n3} are almost identical to those of the initial model in Figure \ref{n4}, proving that this method has little impact. As shown in Figure \ref{n4}, we use the logit of zebra as the loss to calculate gradients and update attention maps. With the update of attention maps, the model's positive attention to zebra gradually shifts to elephant, and the attention to elephant remains positive. After the attention to zebra completely shifts to the elephant area, the judgment area of the two overlaps. The same applies to the second picture of cat and dog in Figure \ref{n4}.

\subsubsection{Single Category Image Interpretation}
\begin{figure}[]
    \begin{center}
    \centerline{\includegraphics[width=0.6\columnwidth]{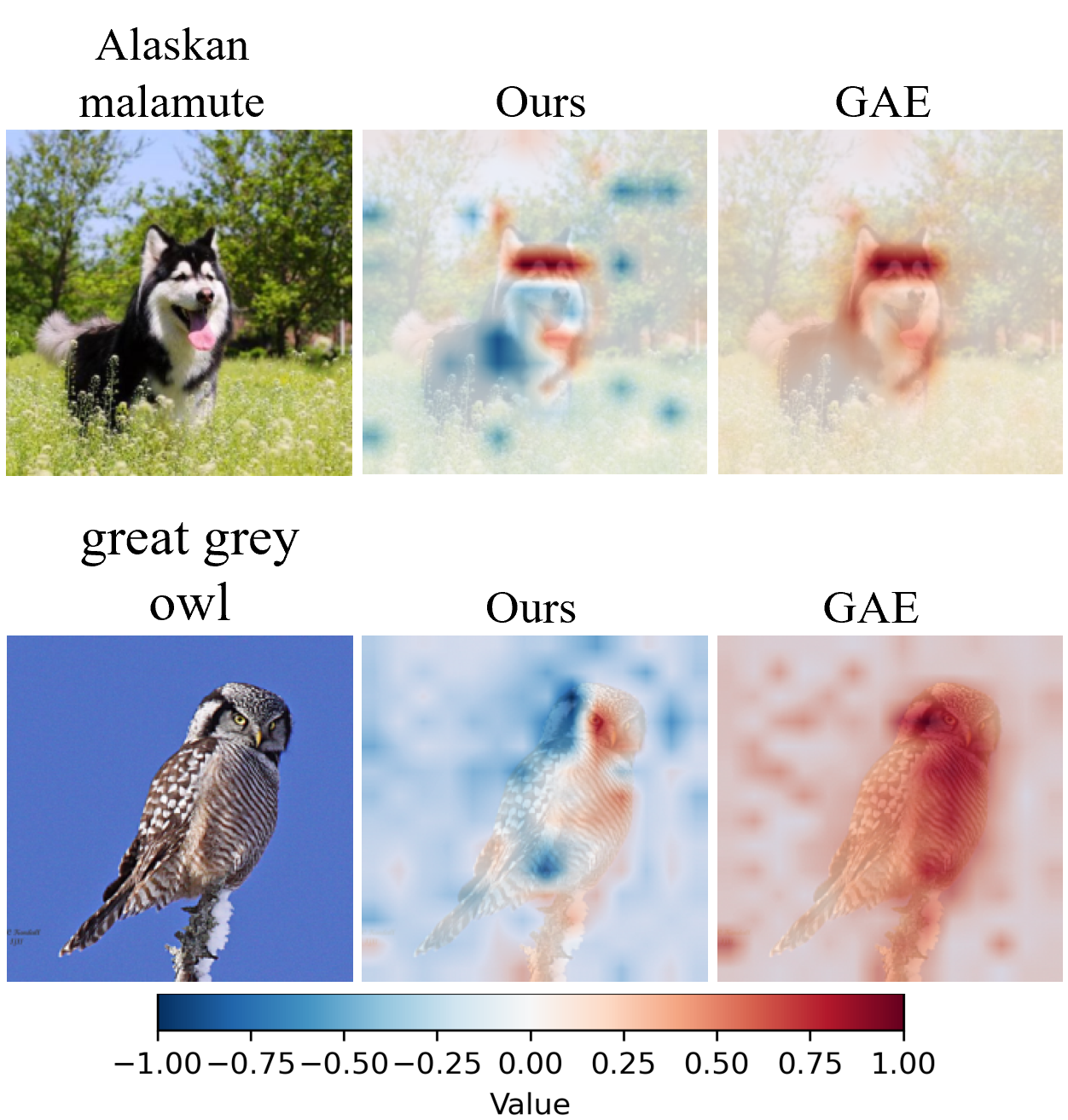}}
    \vspace{-3mm}
    \caption{Single category image interpretation.(Complete gradient correction method cannot interpret single category image directly.)}
    \label{n5}
    \end{center}
    \vspace{-0.8cm}
    \end{figure}

First, let's explain why complete gradient correction performs poorly when interpreting single-class images as shown in Figure \ref{n1} (b) and Figure \ref{n5}. As we explained earlier regarding the model's mechanism, Transformer focuses solely on learned class features while ignoring the background. When there is only one class in the image, all 1000 classes (taking ViT as an example) of the model rely on the same feature region for inference. The model's attention can only transfer within the region of this single class. Moreover, the starting point of attention transfer is the target category. However, the direction of the transfer is not dominated by another category as in multi-class images; instead, it is collectively guided by the other 999 categories, ultimately converging into a direction that is difficult to interpret, leading to the poor interpretation results. This synthesis direction even points towards the background.
\begin{figure}[]
    \begin{center}
    \centerline{\includegraphics[width=0.6\columnwidth]{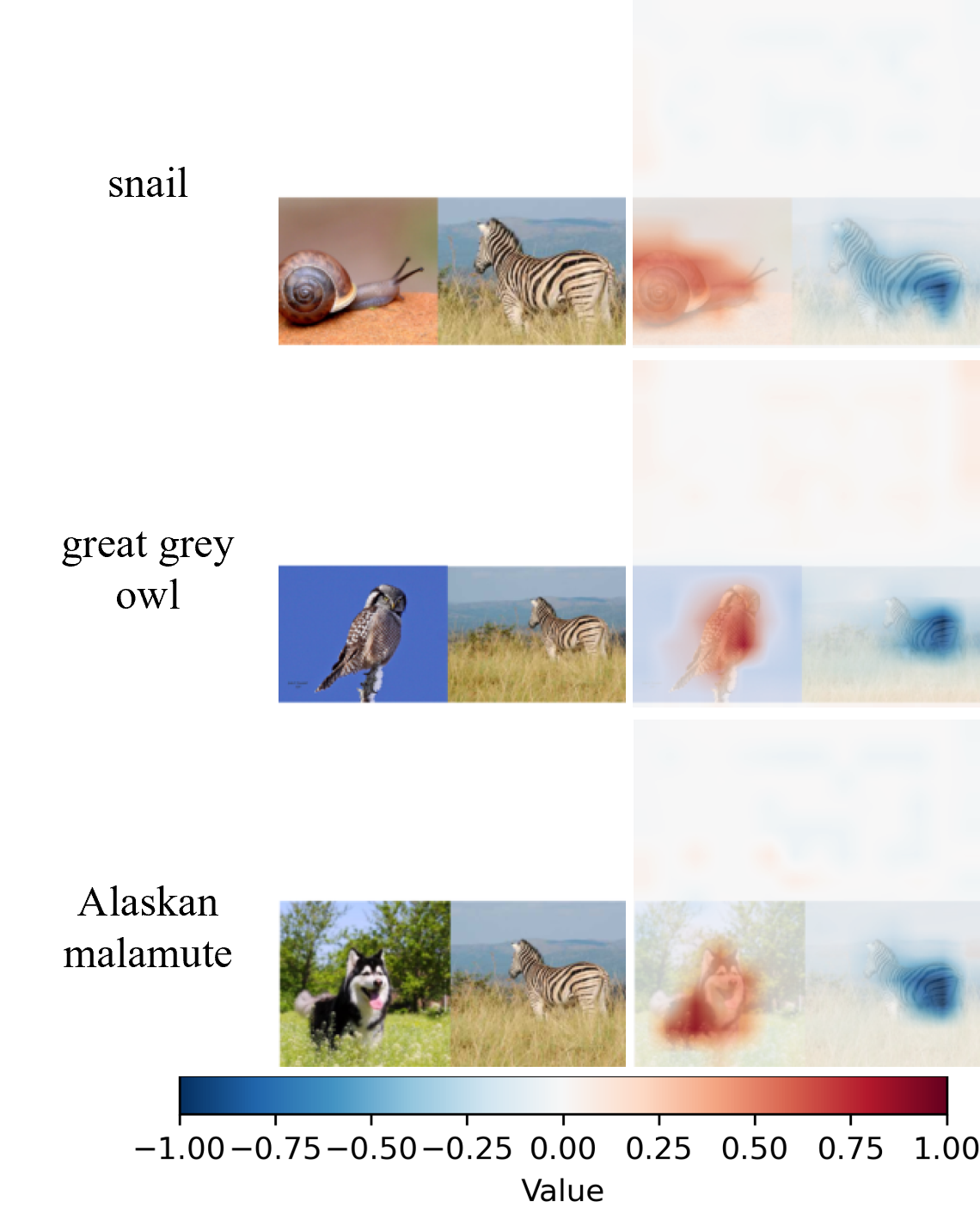}}
    \vspace{-3mm}
    \caption{Guide attention transfer direction to interpret ViT in single category image recognition task.}
    \label{n6}
    \end{center}
    \vspace{-0.8cm}
    \end{figure}
So, we need to divert the model's attention away from the target feature region. This requires the involvement of another class. As shown in Figure \ref{n6}, when there is another class to guide the attention, this method can fully interpret the target feature region. Of course, this formally transforms a single-class image into a multi-class image.

\subsubsection{Detail Interpretation}\label{detail}
Detail interpretation is targeted at single-category images because other classes can interfere with the direction of attention transfer in multi-category images. 

\begin{figure}[]
    \begin{center}
    \centerline{\includegraphics[width=1\columnwidth]{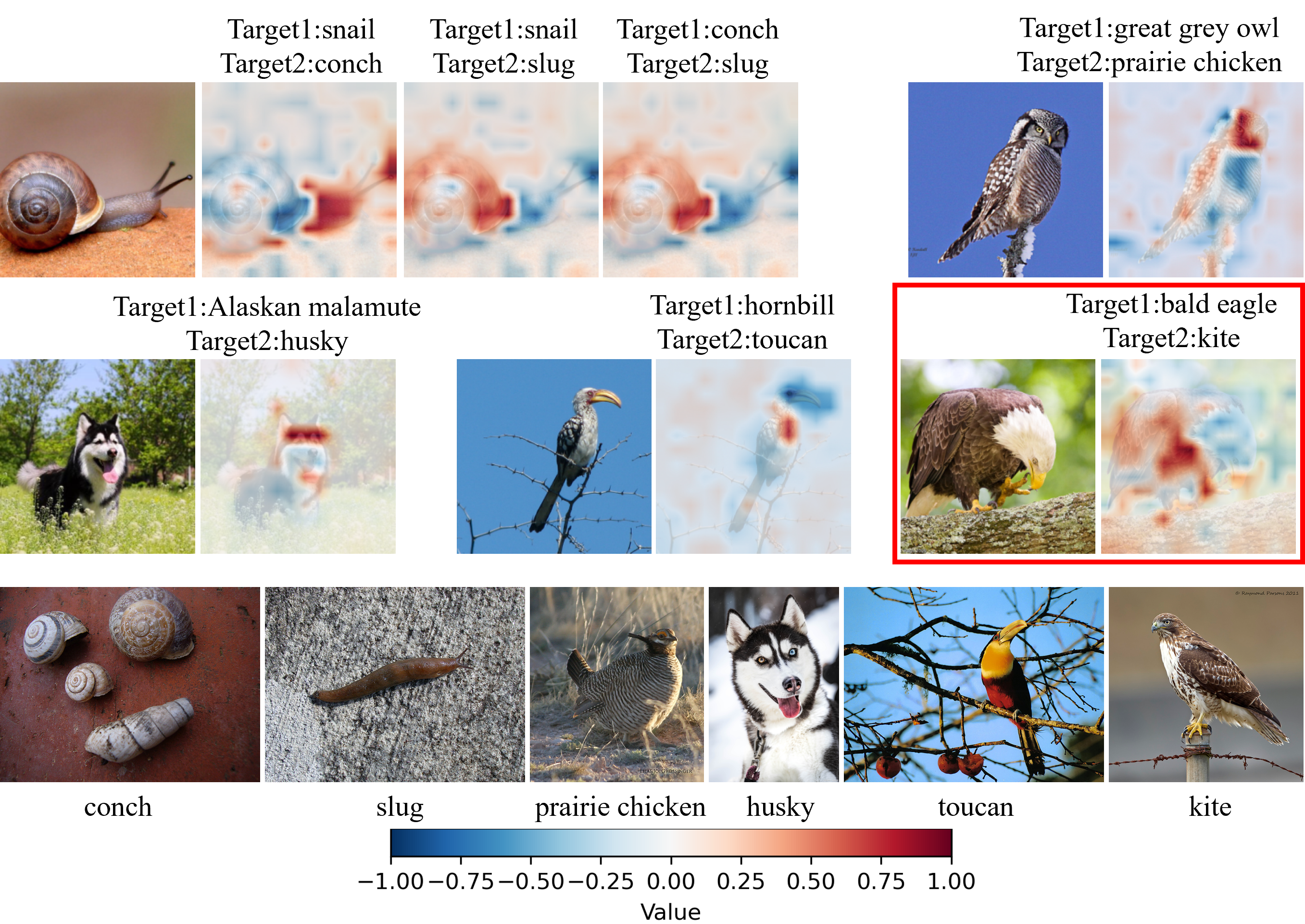}}
    \vspace{-3mm}
    \caption{Detail interpretation.(The last row shows typical images of $target2$.)}
    \label{n7}
    \end{center}
    \vspace{-0.8cm}
    \end{figure}

In the previous subsection, we guided the attention direction outside the target feature region. Here, we perform attention guidance within the target feature region to achieve detail interpretation. We define detail interpretation as interpreting which detail features the model relies on to infer a certain class. For instance, taking the snail in Figure \ref{n1} (b) as an example, the top three outputs of the model are snail, conch, and slug. We aim to interpret which features the model uses to infer the snail, which for the conch, and which for the slug, as shown in Figure \ref{n7}.

To guide attention, we use the loss in Equation \ref{eq8}. The loss can also be $loss = logit_1 - logit_2$, or $loss = logit_1 / logit_2$. Any form of loss function is acceptable as long as it can effectively guide the direction of attention; in other words, any loss function that reduces the logit of $target1$ while increasing the logit of $target2$ effectively would suffice.

The results of the detail interpretation reflect which features are more inclined to lead the model to infer $target1$ (red regions) versus those that predispose the model toward $target2$ predictions, when the model simultaneously accounts for both classes within the loss function.

\textbf{Here's a practical application.}

\begin{figure}[]
    \begin{center}
    \centerline{\includegraphics[width=1\columnwidth]{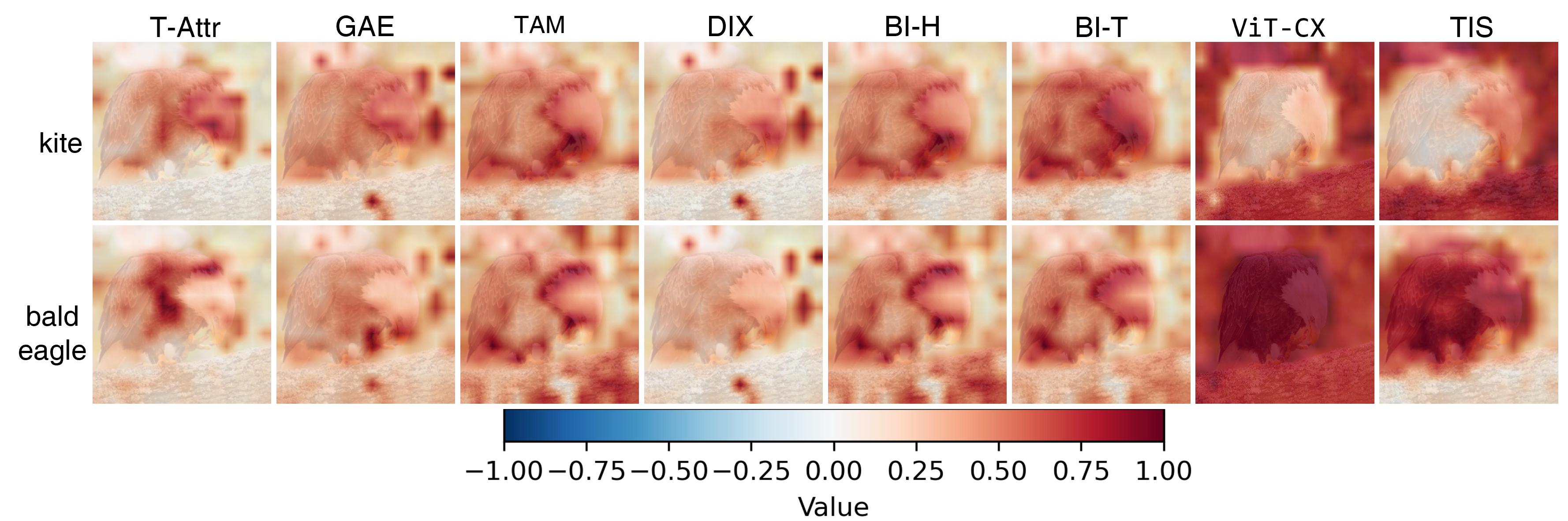}}
    \vspace{-3mm}
    \caption{Interpretations provided by other methods for the bald eagle case.}
    \label{xijiebuchong}
    \end{center}
    \vspace{-0.8cm}
    \end{figure}

The case of the bald eagle comes from TIS \cite{DBLP:conf/iccvw/EnglebertSNMSCV23} paper. Its label is "kite", and the model identifies it as "bald eagle". The experimental results are presented in two locations: the detail interpretation of our method is indicated within the red-framed annotations in Figure \ref{n7}, while the interpretative results of other methods are shown in Figure \ref{xijiebuchong}.

TIS concluded through its analysis of the image that "When generating the saliency maps for both the target class from the ImageNet Dataset and the model predicted class, we noticed that major disagreements between the ground truth and the model can lead to bad saliency maps for the target class, and good saliency maps for the model predicted class." The proposed solution was that "we discovered that highlighting the target class can be forced by removing the softmax at the end of the model. However, this comes at the price of class specificity." Nevertheless, through our detail interpretation, it can be observed that the model erroneously identifies the most iconic white head feature of the bald eagle as a kite feature, which is clearly abnormal. When we meticulously examined the ILSVRC-2012 ImageNet \cite{62} dataset used for fine-tuning ViT, we discovered that the kite category mistakenly included a substantial number of images of bald eagles. This is the fundamental reason behind the model's misidentification of features.

Only our detail interpretation can uncover this issue, while other methods are unable to do so. This underscores the practical value of detail interpretation.

It can be observed that in Figure \ref{n1} (b) and Figure \ref{n2}, when the loss is set as the snail's logit, the interpretation results consistently highlight the connection between the snail's head and its shell, regardless of the correction method employed. This is because this area represents the most distinct feature differentiating snails from conchs and slugs. However, when we direct the attention away from the snail region, this phenomenon does not occur, as illustrated in Figure \ref{n6}. This is also a manifestation of detail interpretation.

\subsection{Rewrite Image Category}\label{rewrite}
The aforementioned research findings reveal that the way Transformer focuses its attention on target regions closely resembles how humans perceive images. However, currently, apart from ViT-CX, which achieves model interpretability by perturbing the original image, other algorithms all provide interpretation at the patch level. The aforementioned research content also operates at the patch level, and we are still unable to achieve pixel-level interpretation. This limitation stems from the fact that the ViT model perceives images through attention mechanisms on a patch-by-patch basis, a method that differs from how humans perceive images. If we leverage this discrepancy and update the image from the model's perspective, the updated image, when viewed from a human perspective, may yield different results.

\begin{figure}[]
    \begin{center}
    \centerline{\includegraphics[width=0.6\columnwidth]{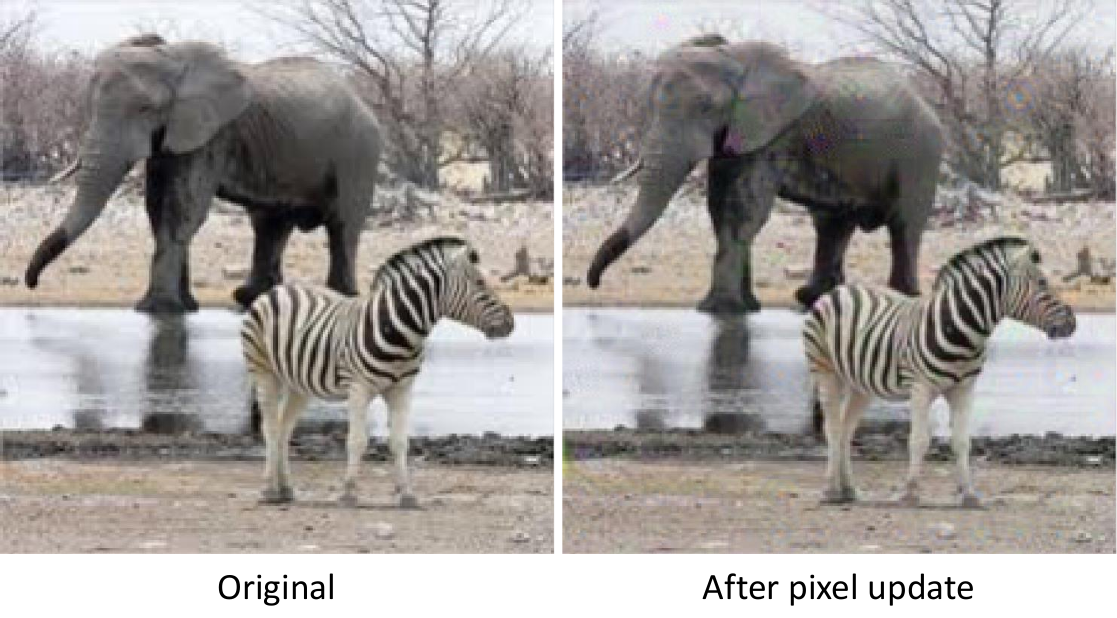}}
    \vspace{-3mm}
    \caption{Pixel update experiment. (We set $loss = logit_{tusker} - logit_{bull mastiff}$. The top-1 output for the original image is ``tusker" with a probability of 35.1\% and a logit of 11.150, while the top-1 output for the updated image is ``bull mastiff" with a probability of 93.8\% and a logit of 12.926.)}
    \label{rewritefig}
    \end{center}
    \vspace{-0.8cm}
    \end{figure}

In Figure \ref{rewritefig}, we attempt to update the image pixels with their gradients, rewriting its content to align with a category we have set. As shown in the figure, we have rewritten the content of the image from the model's perspective, yet there is almost no noticeable change from a human's viewpoint.

The experiment in Figure \ref{rewritefig} is an extension based on our understanding of Transformers from previous sections and is not the focal point of this paper. Although we are uncertain about the specific scenarios, this phenomenon may pose security risks in certain situations.

\subsection{Comparative Experiments}
\subsubsection{Quantitative Experiments}

We reiterate once again that this paper is a study on the characteristics of gradient and attention guidance in the interpretation of Transformer, aiming to fill the gap in the industry's neglect of this aspect. This study is the first to achieve the simultaneous allocation of both positive and negative attention, while also enabling detail interpretation. Given the objectives of our research, we could have initially omitted a comparative analysis through perturbation experiments. However, some readers might expect such an experiment as a matter of convention, so we have included perturbation experiments here. Since perturbation experiments demand substantial computational resources and our method shares the closest architectural similarity with GAE, we only conduct a comparison with GAE in this context.

In the experiment, we use the validation set of ImageNet \cite{62} (ILSVRC) 2012, sample 5000 images evenly from it, and use the zebra images in Figure \ref{n6} to stitch the images together to ensure that each image has a guiding category. We exclude zebra images from the 5000 selected images because they belong to the same category as the guide image. In the experiment, the object we interpret is still the label category of the original image. In the positive perturbation test, tokens are masked from the highest score (most relevant) to the lowest, and lower area-under-the-curve (AUC) is better, indicating that the relevance does reflect their importance to the classification score. And vice versa in the negative perturbation test.

\begin{figure}
    \begin{center}
    \centerline{\includegraphics[width=0.8\columnwidth]{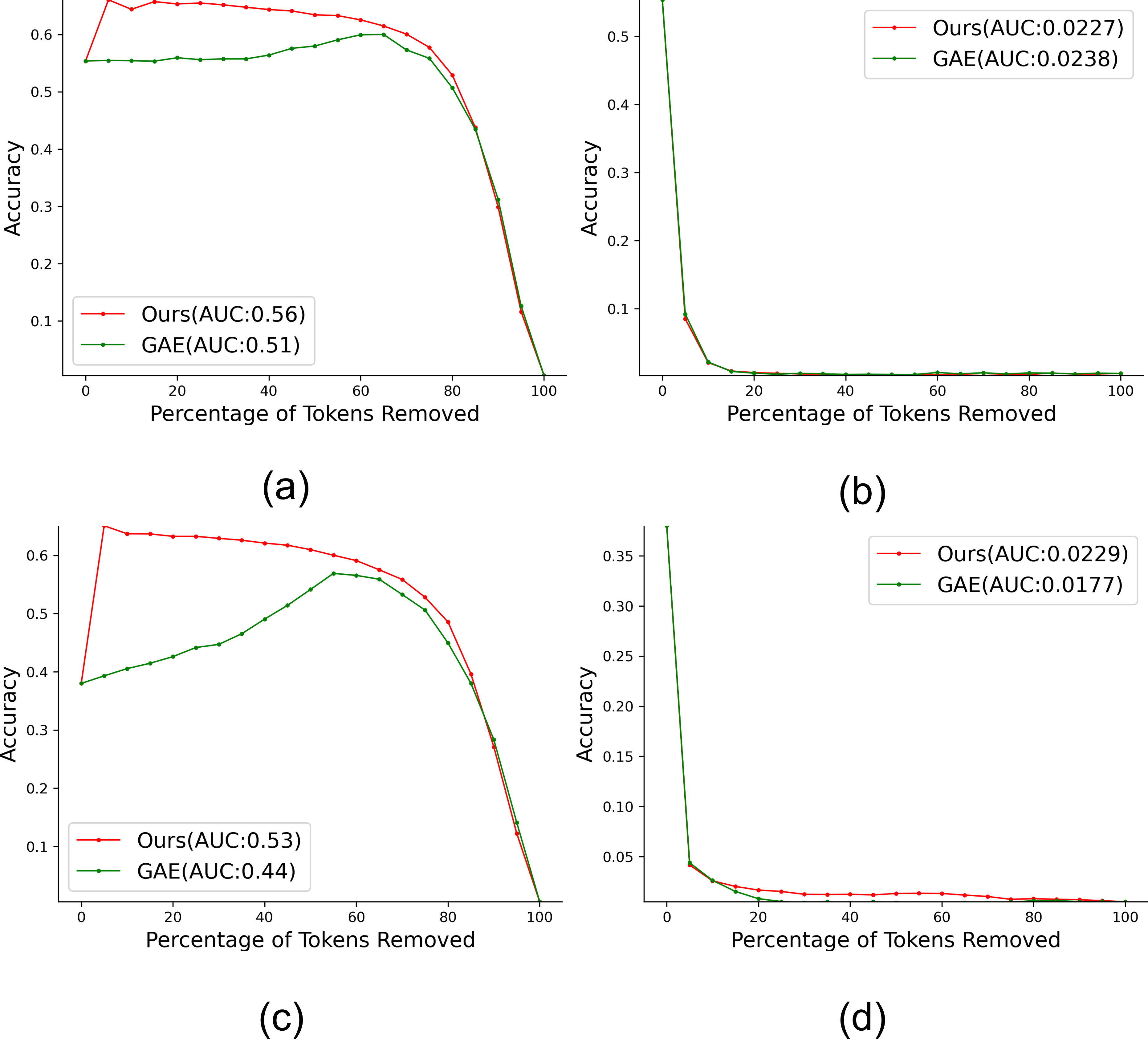}}
    \vspace{-3mm}
    \caption{Perturbation experiments.(a)ViT\_base negative perturbation result.(b)ViT\_base positive perturbation result.(c)ViT\_large negative perturbation result.(d)ViT\_large positive perturbation result.}
    \label{dingliang}
    \end{center}
    \vspace{-0.8cm}
    \end{figure}

The result is shown in Figure \ref{dingliang}. In the positive perturbation test, the effects of the two methods are very similar. In the negative perturbation test, our method is clearly better than GAE. Because our method can assign negative attention to opposite category (zebra), which are the largest perturbations to the target category, and will be masked first according to our method. Therefore, at the beginning, the recognition rate of the target category increased.

It should be noted that perturbation experiments serve as a crude indicator and cannot reflect the amount of detailed noise as accurately as qualitative experiments. For instance, GAE assigns positive contributions to the background and opposing classes, but as long as these positive contributions are smaller than those in the target class region, the positive perturbation experiment will not reveal any differences.

\subsubsection{Qualitative Experiments}
The primary focus of this paper is to investigate the characteristics of gradient and attention guidance in the interpretation of Transformers. The method we propose leverages both gradient and attention guidance to simultaneously assign positive and negative contributions and enable detail interpretation. To the best of our knowledge, this is the first time such an approach has been proposed, therefore we cannot compare with other algorithms in these two dimensions. 

Here, we conduct a comparative analysis of transformer interpretation methods capable of distinguishing between categories in the task of interpreting multi-class images. We exclude attention rollout and Grad-ECLIP from our comparison because attention rollout cannot differentiate between categories, and Grad-ECLIP cannot be applied to the interpretation of multi-head attention.

\begin{figure}[]
    \begin{center}
    \centerline{\includegraphics[width=1\columnwidth]{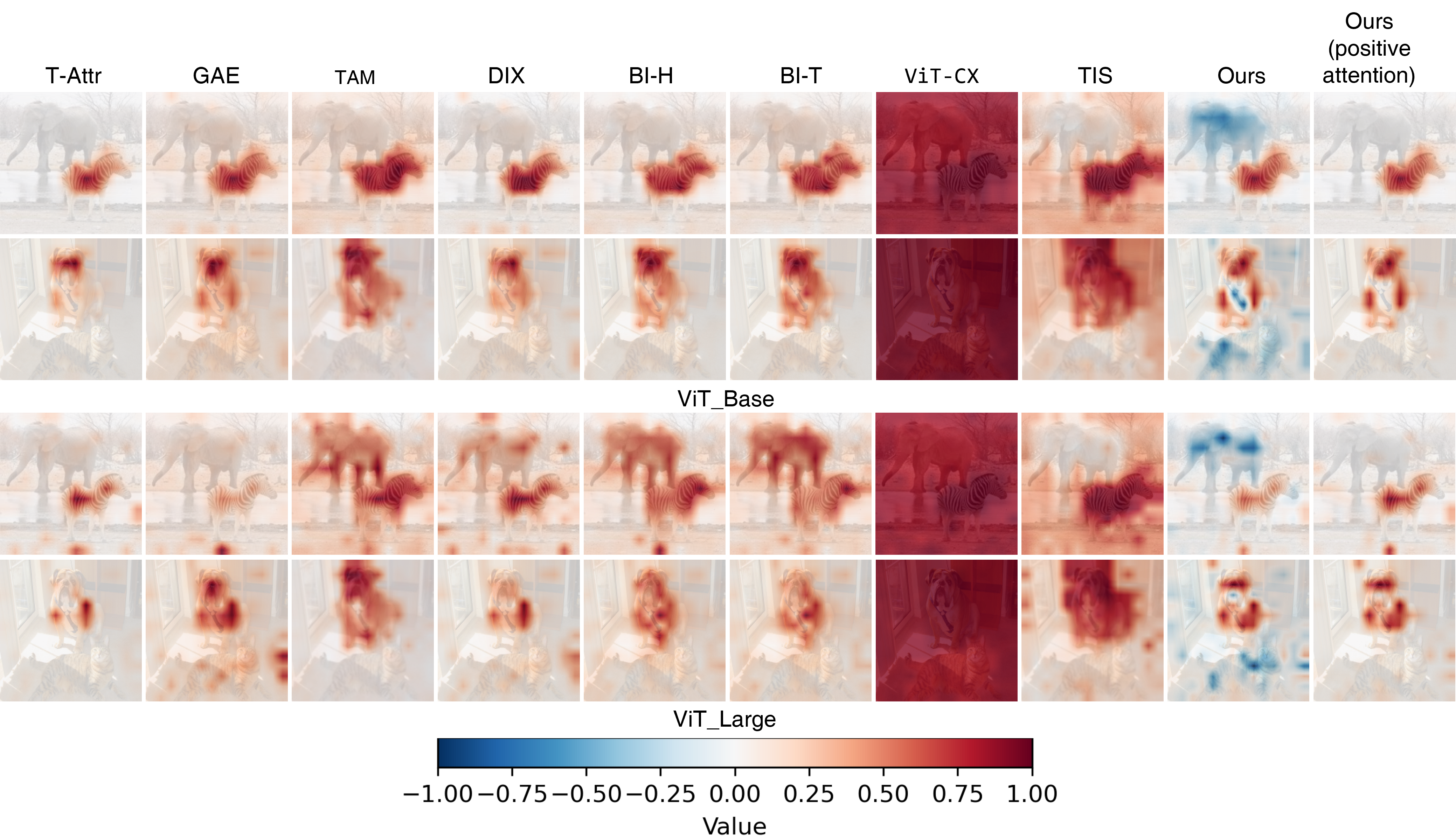}}
    \vspace{-3mm}
    \caption{Interpretation results of ViT\_Base and ViT\_Large in multi category image recognition task.(The categories to be interpreted, listed from top to bottom, are Zebra, Bull mastiff, Zebra, Bull mastiff.)}
    \label{duibi}
    \end{center}
    \vspace{-0.8cm}
    \end{figure}

In Figure \ref{duibi}, it can be found that only our method can allocate both positive and negative contributions simultaneously, thus avoiding the drawbacks of other methods that allocate positive contributions to opposing categories. Moreover, it can be found that some algorithms have difficulty in distinguishing categories.

\section{Conclusion}\label{conclusion}

Gradient and attention guidance play an important role in the interpretation of the Transformer. This paper has conducted an in-depth research on this topic, achieving the simultaneous allocation of negative and positive attention, implementing detail interpretation, and using the properties of gradients to rewrite the categories of images. These achievements help to better understand the working mechanism of Transformer and more accurately test whether the model has learned real knowledge.



\bibliographystyle{named}
\bibliography{danmotai}

\begin{thebibliography}{}

\bibitem[\protect\citeauthoryear{43}{2019}]{43}
European commission. {Ethics} guidelines for trustworthy ai.
\newblock 2019.

\bibitem[\protect\citeauthoryear{44}{2023}]{44}
Select committee on artificial intelligence of the national science and
  technology council. {National} artificial intelligence research and
  development strategic plan.
\newblock 2023.

\bibitem[\protect\citeauthoryear{Abnar and Zuidema}{2020}]{58}
Samira Abnar and Willem~H. Zuidema.
\newblock Quantifying attention flow in transformers.
\newblock In Dan Jurafsky, Joyce Chai, Natalie Schluter, and Joel~R. Tetreault,
  editors, {\em Proceedings of the 58th Annual Meeting of the Association for
  Computational Linguistics, {ACL} 2020, Online, July 5-10, 2020}, pages
  4190--4197. Association for Computational Linguistics, 2020.

\bibitem[\protect\citeauthoryear{Barkan \bgroup \em et al.\egroup }{2023}]{47}
Oren Barkan, Yehonatan Elisha, Jonathan Weill, Yuval Asher, Amit Eshel, and
  Noam Koenigstein.
\newblock Deep integrated explanations.
\newblock In Ingo Frommholz, Frank Hopfgartner, Mark Lee, Michael Oakes, Mounia
  Lalmas, Min Zhang, and Rodrygo L.~T. Santos, editors, {\em Proceedings of the
  32nd {ACM} International Conference on Information and Knowledge Management,
  {CIKM} 2023, Birmingham, United Kingdom, October 21-25, 2023}, pages 57--67.
  {ACM}, 2023.

\bibitem[\protect\citeauthoryear{Binder \bgroup \em et al.\egroup }{2016}]{53}
Alexander Binder, Gr{\'{e}}goire Montavon, Sebastian Lapuschkin, Klaus{-}Robert
  M{\"{u}}ller, and Wojciech Samek.
\newblock Layer-wise relevance propagation for neural networks with local
  renormalization layers.
\newblock In Alessandro E.~P. Villa, Paolo Masulli, and Antonio Javier~Pons
  Rivero, editors, {\em Artificial Neural Networks and Machine Learning -
  {ICANN} 2016 - 25th International Conference on Artificial Neural Networks,
  Barcelona, Spain, September 6-9, 2016, Proceedings, Part {II}}, volume 9887
  of {\em Lecture Notes in Computer Science}, pages 63--71. Springer, 2016.

\bibitem[\protect\citeauthoryear{Chefer \bgroup \em et al.\egroup }{2021a}]{48}
Hila Chefer, Shir Gur, and Lior Wolf.
\newblock Generic attention-model explainability for interpreting bi-modal and
  encoder-decoder transformers.
\newblock In {\em 2021 {IEEE/CVF} International Conference on Computer Vision,
  {ICCV} 2021, Montreal, QC, Canada, October 10-17, 2021}, pages 387--396.
  {IEEE}, 2021.

\bibitem[\protect\citeauthoryear{Chefer \bgroup \em et al.\egroup }{2021b}]{49}
Hila Chefer, Shir Gur, and Lior Wolf.
\newblock Transformer interpretability beyond attention visualization.
\newblock In {\em {IEEE} Conference on Computer Vision and Pattern Recognition,
  {CVPR} 2021, virtual, June 19-25, 2021}, pages 782--791. Computer Vision
  Foundation / {IEEE}, 2021.

\bibitem[\protect\citeauthoryear{Chen \bgroup \em et al.\egroup }{2021}]{12}
Xie Chen, Yu~Wu, Zhenghao Wang, Shujie Liu, and Jinyu Li.
\newblock Developing real-time streaming transformer transducer for speech
  recognition on large-scale dataset.
\newblock In {\em {IEEE} International Conference on Acoustics, Speech and
  Signal Processing, {ICASSP} 2021, Toronto, ON, Canada, June 6-11, 2021},
  pages 5904--5908. {IEEE}, 2021.

\bibitem[\protect\citeauthoryear{Chen \bgroup \em et al.\egroup }{2023}]{46}
Jiamin Chen, Xuhong Li, Lei Yu, Dejing Dou, and Haoyi Xiong.
\newblock Beyond intuition: Rethinking token attributions inside transformers.
\newblock {\em Trans. Mach. Learn. Res.}, 2023, 2023.

\bibitem[\protect\citeauthoryear{Cui \bgroup \em et al.\egroup }{2023}]{38}
Yiming Cui, Ziqing Yang, and Xin Yao.
\newblock Efficient and effective text encoding for chinese llama and alpaca.
\newblock {\em CoRR}, abs/2304.08177, 2023.

\bibitem[\protect\citeauthoryear{Dong \bgroup \em et al.\egroup }{2018}]{13}
Linhao Dong, Shuang Xu, and Bo~Xu.
\newblock Speech-transformer: {A} no-recurrence sequence-to-sequence model for
  speech recognition.
\newblock In {\em 2018 {IEEE} International Conference on Acoustics, Speech and
  Signal Processing, {ICASSP} 2018, Calgary, AB, Canada, April 15-20, 2018},
  pages 5884--5888. {IEEE}, 2018.

\bibitem[\protect\citeauthoryear{Dosovitskiy \bgroup \em et al.\egroup
  }{2021}]{6}
Alexey Dosovitskiy, Lucas Beyer, Alexander Kolesnikov, Dirk Weissenborn,
  Xiaohua Zhai, Thomas Unterthiner, Mostafa Dehghani, Matthias Minderer, Georg
  Heigold, Sylvain Gelly, Jakob Uszkoreit, and Neil Houlsby.
\newblock An image is worth 16x16 words: Transformers for image recognition at
  scale.
\newblock In {\em 9th International Conference on Learning Representations,
  {ICLR} 2021, Virtual Event, Austria, May 3-7, 2021}. OpenReview.net, 2021.

\bibitem[\protect\citeauthoryear{Englebert \bgroup \em et al.\egroup
  }{2023}]{DBLP:conf/iccvw/EnglebertSNMSCV23}
Alexandre Englebert, S{\'{e}}drick Stassin, G{\'{e}}raldin Nanfack, Sidi~Ahmed
  Mahmoudi, Xavier Siebert, Olivier Cornu, and Christophe~De Vleeschouwer.
\newblock Explaining through transformer input sampling.
\newblock In {\em {IEEE/CVF} International Conference on Computer Vision,
  {ICCV} 2023 - Workshops, Paris, France, October 2-6, 2023}, pages 806--815.
  {IEEE}, 2023.

\bibitem[\protect\citeauthoryear{Hu \bgroup \em et al.\egroup }{2020}]{21}
Ronghang Hu, Amanpreet Singh, Trevor Darrell, and Marcus Rohrbach.
\newblock Iterative answer prediction with pointer-augmented multimodal
  transformers for textvqa.
\newblock In {\em 2020 {IEEE/CVF} Conference on Computer Vision and Pattern
  Recognition, {CVPR} 2020, Seattle, WA, USA, June 13-19, 2020}, pages
  9989--9999. Computer Vision Foundation / {IEEE}, 2020.

\bibitem[\protect\citeauthoryear{Ji \bgroup \em et al.\egroup }{2021}]{29}
Yuanfeng Ji, Ruimao Zhang, Huijie Wang, Zhen Li, Lingyun Wu, Shaoting Zhang,
  and Ping Luo.
\newblock Multi-compound transformer for accurate biomedical image
  segmentation.
\newblock In Marleen de~Bruijne, Philippe~C. Cattin, St{\'{e}}phane Cotin,
  Nicolas Padoy, Stefanie Speidel, Yefeng Zheng, and Caroline Essert, editors,
  {\em Medical Image Computing and Computer Assisted Intervention - {MICCAI}
  2021 - 24th International Conference, Strasbourg, France, September 27 -
  October 1, 2021, Proceedings, Part {I}}, volume 12901 of {\em Lecture Notes
  in Computer Science}, pages 326--336. Springer, 2021.

\bibitem[\protect\citeauthoryear{Jiang \bgroup \em et al.\egroup }{2021}]{52}
Peng{-}Tao Jiang, Chang{-}Bin Zhang, Qibin Hou, Ming{-}Ming Cheng, and Yunchao
  Wei.
\newblock Layercam: Exploring hierarchical class activation maps for
  localization.
\newblock {\em {IEEE} Trans. Image Process.}, 30:5875--5888, 2021.

\bibitem[\protect\citeauthoryear{Li \bgroup \em et al.\egroup }{2019}]{22}
Liunian~Harold Li, Mark Yatskar, Da~Yin, Cho{-}Jui Hsieh, and Kai{-}Wei Chang.
\newblock Visualbert: {A} simple and performant baseline for vision and
  language.
\newblock {\em CoRR}, abs/1908.03557, 2019.

\bibitem[\protect\citeauthoryear{Li \bgroup \em et al.\egroup }{2021}]{23}
Wei Li, Can Gao, Guocheng Niu, Xinyan Xiao, Hao Liu, Jiachen Liu, Hua Wu, and
  Haifeng Wang.
\newblock {UNIMO:} towards unified-modal understanding and generation via
  cross-modal contrastive learning.
\newblock In Chengqing Zong, Fei Xia, Wenjie Li, and Roberto Navigli, editors,
  {\em Proceedings of the 59th Annual Meeting of the Association for
  Computational Linguistics and the 11th International Joint Conference on
  Natural Language Processing, {ACL/IJCNLP} 2021, (Volume 1: Long Papers),
  Virtual Event, August 1-6, 2021}, pages 2592--2607. Association for
  Computational Linguistics, 2021.

\bibitem[\protect\citeauthoryear{Liu \bgroup \em et al.\egroup }{2021}]{7}
Ze~Liu, Yutong Lin, Yue Cao, Han Hu, Yixuan Wei, Zheng Zhang, Stephen Lin, and
  Baining Guo.
\newblock Swin transformer: Hierarchical vision transformer using shifted
  windows.
\newblock In {\em 2021 {IEEE/CVF} International Conference on Computer Vision,
  {ICCV} 2021, Montreal, QC, Canada, October 10-17, 2021}, pages 9992--10002.
  {IEEE}, 2021.

\bibitem[\protect\citeauthoryear{Michel \bgroup \em et al.\egroup }{2019}]{57}
Paul Michel, Omer Levy, and Graham Neubig.
\newblock Are sixteen heads really better than one?
\newblock In Hanna~M. Wallach, Hugo Larochelle, Alina Beygelzimer, Florence
  d'Alch{\'{e}}{-}Buc, Emily~B. Fox, and Roman Garnett, editors, {\em Advances
  in Neural Information Processing Systems 32: Annual Conference on Neural
  Information Processing Systems 2019, NeurIPS 2019, December 8-14, 2019,
  Vancouver, BC, Canada}, pages 14014--14024, 2019.

\bibitem[\protect\citeauthoryear{Russakovsky \bgroup \em et al.\egroup
  }{2015}]{62}
Olga Russakovsky, Jia Deng, Hao Su, Jonathan Krause, Sanjeev Satheesh, Sean Ma,
  Zhiheng Huang, Andrej Karpathy, Aditya Khosla, Michael~S. Bernstein,
  Alexander~C. Berg, and Li~Fei{-}Fei.
\newblock Imagenet large scale visual recognition challenge.
\newblock {\em Int. J. Comput. Vis.}, 115(3):211--252, 2015.

\bibitem[\protect\citeauthoryear{Selvaraju \bgroup \em et al.\egroup
  }{2017}]{51}
Ramprasaath~R. Selvaraju, Michael Cogswell, Abhishek Das, Ramakrishna Vedantam,
  Devi Parikh, and Dhruv Batra.
\newblock Grad-cam: Visual explanations from deep networks via gradient-based
  localization.
\newblock In {\em {IEEE} International Conference on Computer Vision, {ICCV}
  2017, Venice, Italy, October 22-29, 2017}, pages 618--626. {IEEE} Computer
  Society, 2017.

\bibitem[\protect\citeauthoryear{Touvron \bgroup \em et al.\egroup
  }{2023a}]{41}
Hugo Touvron, Thibaut Lavril, Gautier Izacard, Xavier Martinet, Marie{-}Anne
  Lachaux, Timoth{\'{e}}e Lacroix, Baptiste Rozi{\`{e}}re, Naman Goyal, Eric
  Hambro, Faisal Azhar, Aur{\'{e}}lien Rodriguez, Armand Joulin, Edouard Grave,
  and Guillaume Lample.
\newblock Llama: Open and efficient foundation language models.
\newblock {\em CoRR}, abs/2302.13971, 2023.

\bibitem[\protect\citeauthoryear{Touvron \bgroup \em et al.\egroup
  }{2023b}]{42}
Hugo Touvron, Louis Martin, Kevin Stone, Peter Albert, Amjad Almahairi, Yasmine
  Babaei, Nikolay Bashlykov, Soumya Batra, Prajjwal Bhargava, Shruti Bhosale,
  Dan Bikel, Lukas Blecher, Cristian Canton{-}Ferrer, Moya Chen, Guillem
  Cucurull, David Esiobu, Jude Fernandes, Jeremy Fu, Wenyin Fu, Brian Fuller,
  Cynthia Gao, Vedanuj Goswami, Naman Goyal, Anthony Hartshorn, Saghar
  Hosseini, Rui Hou, Hakan Inan, Marcin Kardas, Viktor Kerkez, Madian Khabsa,
  Isabel Kloumann, Artem Korenev, Punit~Singh Koura, Marie{-}Anne Lachaux,
  Thibaut Lavril, Jenya Lee, Diana Liskovich, Yinghai Lu, Yuning Mao, Xavier
  Martinet, Todor Mihaylov, Pushkar Mishra, Igor Molybog, Yixin Nie, Andrew
  Poulton, Jeremy Reizenstein, Rashi Rungta, Kalyan Saladi, Alan Schelten, Ruan
  Silva, Eric~Michael Smith, Ranjan Subramanian, Xiaoqing~Ellen Tan, Binh Tang,
  Ross Taylor, Adina Williams, Jian~Xiang Kuan, Puxin Xu, Zheng Yan, Iliyan
  Zarov, Yuchen Zhang, Angela Fan, Melanie Kambadur, Sharan Narang,
  Aur{\'{e}}lien Rodriguez, Robert Stojnic, Sergey Edunov, and Thomas Scialom.
\newblock Llama 2: Open foundation and fine-tuned chat models.
\newblock {\em CoRR}, abs/2307.09288, 2023.

\bibitem[\protect\citeauthoryear{Vaswani \bgroup \em et al.\egroup }{2017}]{1}
Ashish Vaswani, Noam Shazeer, Niki Parmar, Jakob Uszkoreit, Llion Jones,
  Aidan~N. Gomez, Lukasz Kaiser, and Illia Polosukhin.
\newblock Attention is all you need.
\newblock In Isabelle Guyon, Ulrike von Luxburg, Samy Bengio, Hanna~M. Wallach,
  Rob Fergus, S.~V.~N. Vishwanathan, and Roman Garnett, editors, {\em Advances
  in Neural Information Processing Systems 30: Annual Conference on Neural
  Information Processing Systems 2017, December 4-9, 2017, Long Beach, CA,
  {USA}}, pages 5998--6008, 2017.

\bibitem[\protect\citeauthoryear{Wang \bgroup \em et al.\egroup }{2021}]{30}
Tao Wang, Zhihui Lai, and Heng Kong.
\newblock Tfnet: Transformer fusion network for ultrasound image segmentation.
\newblock In Christian Wallraven, Qingshan Liu, and Hajime Nagahara, editors,
  {\em Pattern Recognition - 6th Asian Conference, {ACPR} 2021, Jeju Island,
  South Korea, November 9-12, 2021, Revised Selected Papers, Part {I}}, volume
  13188 of {\em Lecture Notes in Computer Science}, pages 314--325. Springer,
  2021.

\bibitem[\protect\citeauthoryear{Xie \bgroup \em et al.\egroup }{2021}]{28}
Yutong Xie, Jianpeng Zhang, Chunhua Shen, and Yong Xia.
\newblock Cotr: Efficiently bridging {CNN} and transformer for 3d medical image
  segmentation.
\newblock In Marleen de~Bruijne, Philippe~C. Cattin, St{\'{e}}phane Cotin,
  Nicolas Padoy, Stefanie Speidel, Yefeng Zheng, and Caroline Essert, editors,
  {\em Medical Image Computing and Computer Assisted Intervention - {MICCAI}
  2021 - 24th International Conference, Strasbourg, France, September 27 -
  October 1, 2021, Proceedings, Part {III}}, volume 12903 of {\em Lecture Notes
  in Computer Science}, pages 171--180. Springer, 2021.

\bibitem[\protect\citeauthoryear{Xie \bgroup \em et al.\egroup
  }{2023}]{DBLP:conf/ijcai/Xie0CZ23}
Weiyan Xie, Xiao{-}Hui Li, Caleb~Chen Cao, and Nevin~L. Zhang.
\newblock Vit-cx: Causal explanation of vision transformers.
\newblock In {\em Proceedings of the Thirty-Second International Joint
  Conference on Artificial Intelligence, {IJCAI} 2023, 19th-25th August 2023,
  Macao, SAR, China}, pages 1569--1577. ijcai.org, 2023.

\bibitem[\protect\citeauthoryear{Yuan \bgroup \em et al.\egroup
  }{2021}]{yuan2021explaining}
Tingyi Yuan, Xuhong Li, Haoyi Xiong, Hui Cao, and Dejing Dou.
\newblock Explaining information flow inside vision transformers using markov
  chain.
\newblock In {\em eXplainable AI approaches for debugging and diagnosis.},
  2021.

\bibitem[\protect\citeauthoryear{Zhao \bgroup \em et al.\egroup
  }{2024}]{DBLP:conf/icml/ZhaoWZZC24}
Chenyang Zhao, Kun Wang, Xingyu Zeng, Rui Zhao, and Antoni~B. Chan.
\newblock Gradient-based visual explanation for transformer-based {CLIP}.
\newblock In {\em Forty-first International Conference on Machine Learning,
  {ICML} 2024, Vienna, Austria, July 21-27, 2024}. OpenReview.net, 2024.

\end{thebibliography}

\end{document}